\documentclass[lettersize,journal]{IEEEtran}
\usepackage{amsmath,amsfonts}
\usepackage{algorithmic}
\usepackage{algorithm}
\usepackage{array}
\usepackage[caption=false,font=normalsize,labelfont=sf,textfont=sf]{subfig}
\usepackage{textcomp}
\usepackage{stfloats}
\usepackage{url}
\usepackage{verbatim}
\usepackage{graphicx}
\usepackage{cite}
\usepackage{multirow}
\usepackage{threeparttable}
\usepackage{booktabs}
\usepackage[normalem]{ulem}

\usepackage{array}
\usepackage{threeparttable}
\usepackage[dvipsnames]{xcolor}
\usepackage[accsupp]{axessibility}

\usepackage{pifont}
\newcommand{\xmark}{\ding{55}}%
\newcommand{\cmark}{\ding{51}}%

\newcommand{\Hl}[2][\empty]{%
\ifx#1\empty
\else
\sethlcolor{#1}%
\fi
\hl{#2}}
\usepackage{soul,color}
\soulregister\Hl{7}
\soulregister\ref7
\soulregister\cite7
\soulregister\pageref7
\makeatletter 
\newcount\SOUL@minus
\makeatother  

\newcommand*\rot{\rotatebox{90}}
\def\ourapproach{SiNC }

\definecolor{myorange}{RGB}{220, 112, 3}
\definecolor{mygreen}{RGB}{47, 154, 47}
\definecolor{myblue}{RGB}{115, 60, 160}

\begin{document}

 \title{SiNC+: Adaptive Camera-Based Vitals with Unsupervised Learning of Periodic Signals}

\author{Jeremy Speth, Nathan Vance, Patrick Flynn, Adam Czajka
\thanks{Authors are with the Computer Vision Research Lab (CVRL) at the University of Notre Dame. Email for corresponding author: jspeth@nd.edu}}



\maketitle

\begin{abstract}
Subtle periodic signals, such as blood volume pulse and respiration, can be extracted from RGB video, enabling noncontact health monitoring at low cost.
Advancements in remote pulse estimation -- or remote photoplethysmography (rPPG) -- are currently driven by deep learning solutions.
However, modern approaches are trained and evaluated on benchmark datasets with ground truth from contact-PPG sensors.
We present the first non-contrastive unsupervised learning framework for signal regression to mitigate the need for labelled video data.
With minimal assumptions of periodicity and finite bandwidth, our approach discovers the blood volume pulse directly from unlabelled videos.
We find that encouraging sparse power spectra within normal physiological bandlimits and variance over batches of power spectra is sufficient for learning visual features of periodic signals.
We perform the first experiments utilizing unlabelled video data not specifically created for rPPG to train robust pulse rate estimators.
Given the limited inductive biases, we successfully applied the same approach to camera-based respiration by changing the bandlimits of the target signal.
This shows that the approach is general enough for unsupervised learning of bandlimited quasi-periodic signals from different domains.
Furthermore, we show that the framework is effective for finetuning models on unlabelled video from a single subject, allowing for personalized and adaptive signal regressors. 

\end{abstract}


\section{Introduction}
\label{sec:intro}

\IEEEPARstart{C}{amera-based}
vitals estimation is a rapidly growing field enabling non-contact health monitoring in a variety of settings~\cite{McDuff2022}.
Although many of the signals avoid detection from the human eye, video data in the visible and infrared ranges contain subtle intensity changes caused by physiological oscillations such as blood volume and respiration.
Significant remote photoplethysmography (rPPG) research for estimating the cardiac pulse has leveraged supervised deep learning for robust signal extraction~\cite{Chen2018,Yu2019,Niu2020,Liu_MTTS_2020,Speth_CVIU_2021,Yu_2022_CVPR}.
While the number of successful approaches has rapidly increased, the size of benchmark video datasets with simultaneous vitals recordings has remained relatively stagnant.

Robust deep learning-based systems for deployment require training on larger volumes of video data with diverse skin tones, lighting, camera sensors, and movement.
However, collecting simultaneous video and physiological ground truth with contact-PPG or electrocardiograms (ECG) is challenging for several reasons.
First, many hours of high quality videos is an unwieldy volume of data. Second, recording a diverse subject population in conditions representative of real-world activities is difficult to conduct in the lab setting. Finally, synchronizing contact measurements with video is technically challenging, and even contact measurements used for ground truth contain noise.

\begin{figure*}
    \centering
    \includegraphics[width=\linewidth]{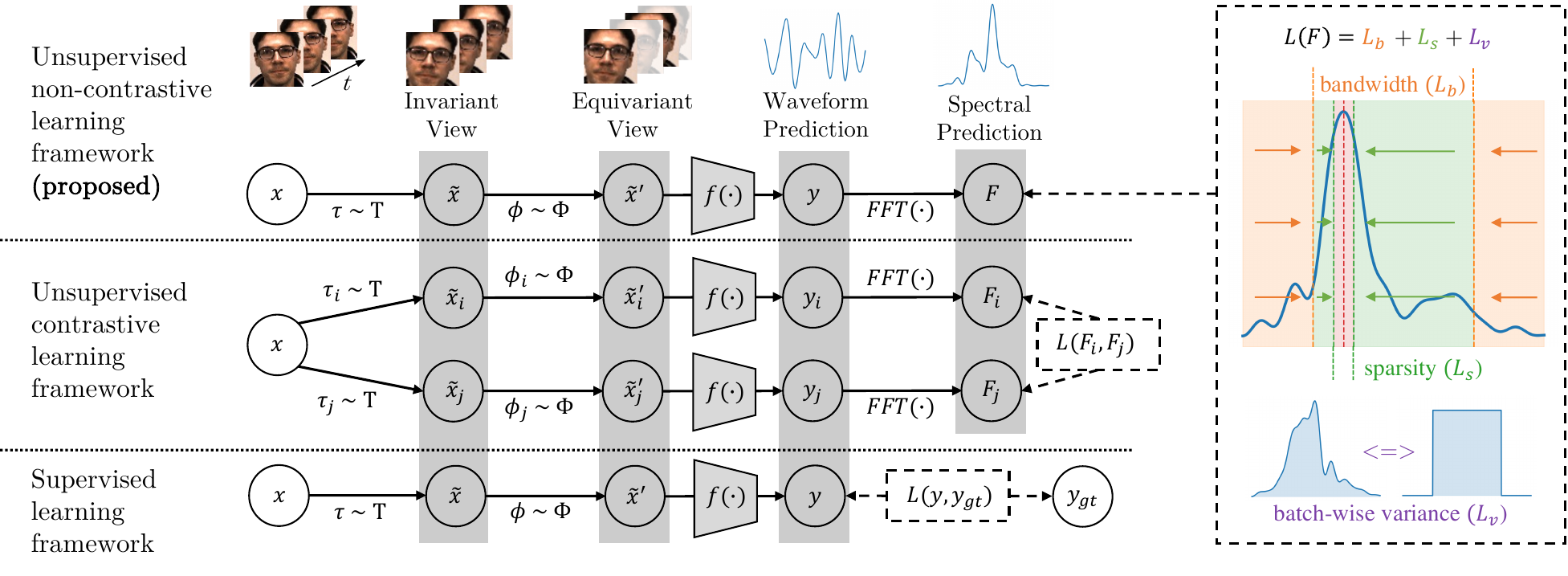}
    \caption{Overview of the \ourapproach framework for rPPG compared with traditional supervised and unsupervised learning. Supervised and contrastive losses use distance metrics to the ground truth or other samples. Our framework applies the loss directly to the prediction by shaping the frequency spectrum, and encouraging \textcolor{myblue}{{\bf variance over a batch}} of inputs. Power outside of the \textcolor{myorange}{{\bf bandlimits}} is penalized to learn invariances to irrelevant frequencies. Power within the bandlimits is encouraged to be \textcolor{mygreen}{{\bf sparsely}} distributed near the peak frequency.}
    \label{fig:framework}
\end{figure*}

Fortunately, recent works find that contrastive unsupervised learning for rPPG is a promising solution to the data scarcity problem~\cite{Gideon_2021_ICCV,Sun_2022_ECCV,Wang_SSL_2022,Yuzhe_SimPer_2022}.
With end-to-end unsupervised learning, collecting more representative training data to learn powerful visual features is much simpler, since only video is required without associated medical information.
However, the contrastive methods do not incorporate prior information on periodic signals into the framework, and typically require a dataset of multiple subjects to form negative pairs.

This paper extends our earlier work on learning
periodic \textbf{si}gnals from video with \textbf{n}on-\textbf{c}ontrastive unsupervised learning (SiNC)~\cite{speth2023sinc}.
In our experiments we found that weak assumptions of periodicity are sufficient for learning the minuscule visual features corresponding to the blood volume pulse from unlabelled face videos.
The loss functions can be computed in the frequency domain over batches without the need for pairwise or triplet comparisons.
Figure \ref{fig:framework} compares the proposed approach with supervised and contrastive unsupervised learning approaches.
In this extended work, we show that the SiNC approach is easily generalized to other domains such as respiratory signals from video by changing the bandlimits in the loss formulation.
Additionally, while most unsupervised deep learning approaches are created with the intention of training on easily gathered large-scale datasets, SiNC can be used for fine-tuning on a single short segment of video from one person.
This opens up new opportunities for privacy-aware, personalized, and adaptive models in remote physiological sensing.

This work creates new opportunities in deep learning for camera-based vitals and learning to estimate quasi-periodic signals from unlabelled data.
The primary contributions of this paper are as follows:
\begin{enumerate}
    \item A general framework for learning periodic \textbf{si}gnals from video with \textbf{n}on-\textbf{c}ontrastive unsupervised learning (SiNC).
    \item Successful application of SiNC to the contactless measurement of pulse and respiration waveforms from video.
    \item The first experiments using unsupervised learning for remote respiration estimation from a single camera.
    \item The first experiments and results of training with non-rPPG-specific video data without ground truth vitals.
    \item Experiments showing that SiNC is effective with very little data for fine-tuning, allowing for personalization and test-time adaptation.
\end{enumerate}

Source code to replicate the original conference work \cite{speth2023sinc} is currently available at \url{https://github.com/CVRL/SiNC-rPPG}. Additional code used in this journal extension will be made available upon acceptance.

\section{Related Work}
\subsection{Remote Photoplethysmography (rPPG)}

Approaches for remote pulse estimation have shifted over the last decade from blind source separation~\cite{Poh2010,Poh2011}, through linear color transformations~\cite{DeHaan2013,DeHaan2014,Wang2017,Wang2019} to training supervised deep learning-based models~\cite{Niu2018,Chen2018,Yu2019,Niu2020,Liu_MTTS_2020,Lee_ECCV_2020,Lu_2021,Speth_CVIU_2021,Zhao_2021,Yu_2022_CVPR}. While the color transformations generalize well across datasets, deep learning-based models give better accuracy when tested on data from a similar distribution to the training set. To this end, much deep learning research has focused on architectures for robust spatial and temporal features from limited datasets.

To get around the data bottleneck, large synthetic datasets have recently been proposed~\cite{Kadambi2022,mcduff2022scamps}. The SCAMPS dataset~\cite{mcduff2022scamps} contains 2,800 videos of synthetic avatars with corresponding PPG, EKG, respiration, and facial action units. The UCLA-synthetic dataset~\cite{Kadambi2022} contains 480 videos, which are shown to improve performance when trained with real data. A strength of synthetic datasets is their ability to cover the broad range of skin tones compared to physical data. Another solution to the lack of physiological data is unsupervised learning, where a large set of videos and periodic priors on the output signal is sufficient~\cite{Gideon_2021_ICCV,Sun_2022_ECCV,Wang_SSL_2022,Yuzhe_SimPer_2022}.

\subsection{Self-Supervised Learning for rPPG}
\label{sec:background_SSL_rPPG}
Self-supervised learning (SSL) is progressing for image representation learning, with two main competing classes of approaches: contrastive and non-contrastive (or regularized) learning~\cite{bardes2022vicreg}. Contrastive approaches~\cite{Misra2019,Caron2020, Chen2020} define criteria for distinguishing whether two or more samples are the same or different. 
Non-contrastive methods augment positive pairs, and enforce variance in the predictions over batches to avoid \textit{collapse}, in which the model's embeddings reside in a small subspace of the feature space~\cite{bardes2022vicreg}. Distillation methods only use positive samples and avoid collapse by applying a moving average and stop-gradient operator~\cite{Grill2020,Chen2021}. Another approach is to maximize information content of embeddings~\cite{Zbontar2021,ermolov2021whitening}.

Applying SSL to rPPG is particularly desirable, due to the challenges of collecting video datasets with pulse ground truth.
Almost all existing unsupervised rPPG approaches are contrastive~\cite{Gideon_2021_ICCV,Sun_2022_ECCV,Wang_SSL_2022,Yuzhe_SimPer_2022,Yue2023,sun2023contrastphys+,akamatsu2023calibrationphys}, in which the same model is fed pairs of input videos, and the predictions over similar videos are pulled closer, while the predictions from dissimilar videos are repelled. Gideon \etal~\cite{Gideon_2021_ICCV} were the first to train a deep learning model without labels for rPPG using the contrastive framework. The core of their approach is frequency resampling to create negative samples. Although spatially similar, a resampled video clip contains a different underlying pulse rate, so the model must learn to attend to the temporal dynamics.
For their distance function between pairs, they calculated the mean square error between power spectral densities of the model's waveform predictions. While their approach learns to estimate the pulse, their formulation with negative samples is imprecise.
The resampled frequency for the negative sample is known, so the relative translation of the power spectrum from the anchor sample can be directly computed. Thus, rather than repelling the estimated spectra, it is more accurate to penalize differences from the known spectra. Furthermore, resampling close to the original sampling rate causes overlap in the power spectra, so repelling the pair is inaccurate.

More recent works incorporated the previously ignored resampling factor into the loss functions~\cite{Yuzhe_SimPer_2022,Yue2023}.
Yuzhe \etal~\cite{Yuzhe_SimPer_2022} modified the InfoNCE loss~\cite{Oord2018} to scale the desired similarity between pairs by their relative sampling rate.
A downside is that their learning framework is not end-to-end unsupervised and requires fine-tuning with PPG labels after the self-supervised stage.

Differently from \cite{Gideon_2021_ICCV}, Contrast-Phys~\cite{Sun_2022_ECCV,sun2023contrastphys+} and SLF-RPM~\cite{Wang_SSL_2022} consider all samples different from the anchor to be negatives. This assumes that the power spectra will vary between subjects or sufficiently long windows for the same subject.
This runs into similar issues with negative pairs as Gideon's approach.
Different subjects may have the same pulse rate, so punishing the model for predicting similar frequencies is common during training.
Furthermore, the Fast Fourier Transform (FFT) does not produce perfectly sparse decompositions, resulting in spectral overlap even if the heart rate differs by several beats per minute (bpm).
As an example, the last column of the second row in Fig. \ref{fig:losses} shows the nulls of the main lobe are nearly 30 bpm apart.

Liu \etal~\cite{Liu2023rPPGMAE} diverge from contrastive learning towards self-supervision with masked autoencoders (rPPG-MAE).
They use spatio-temporal maps from multiple color spaces as the input to a transformer~\cite{Dosovitskiy2021} for their masked reconstruction task during training.
As with \cite{Yuzhe_SimPer_2022} and \cite{Wang_SSL_2022}, rPPG-MAE requires a fine-tuning on labelled videos after pretraining.

To the authors knowledge, Akamatsu \etal~\cite{akamatsu2023calibrationphys} present the first self-supervised learning experiments for remote respiration.
Their approach forms video pairs for contrastive learning from a pair of input videos from different cameras. Positive pairs are the simultaneous videos from different cameras, and negative pairs are videos captured at different times.
While they show efficacy for both pulse and respiration, it requires simultaneous video capture from multiple cameras, which is data-intensive and expensive.

In Sec.~\ref{sec:adaptation}, we show that model personalization and test-time adaptation with our unsupervised learning framework is simple and effective. The most similar related works for rPPG are \cite{Lee_ECCV_2020} and \cite{Liu_2021_Metaphys}. Lee \etal~\cite{Lee_ECCV_2020} use transductive meta-learning for unsupervised adaptation by training a synthetic gradient generator network. The downside of this approach is its complexity, as it requires separate networks for feature extraction and gradient generation. Liu \etal~\cite{Liu_2021_Metaphys} use both supervised and unsupervised meta-learning for rPPG. Their unsupervised approach uses POS~\cite{Wang2017} signals as the targets when personalizing a model. While this may work for some benchmark rPPG datasets where POS has clean signals, it likely fails on datasets with strong motion such as DDPM~\cite{Speth_IJCB_2021,Vance2022}.
\section{Method}

We first formulate signal regression from video. A video sample $x_i \in \mathbb{R}^{T\times W\times H\times C}$ sampled from a dataset $\mathcal{D}$ consists of $T$ images of size $W\times H$ pixels across $C$ channels, captured over time.
State-of-the-art methods offer models $f$ that regress a waveform $\mathbb{R}^T \ni y_i = f(x_i)$ of the same length as the video.
Recently, the task has been effectively modeled end-to-end with the models $f$ being spatiotemporal neural networks~\cite{Yu2019,Lee_ECCV_2020,Speth_CVIU_2021,Lu_2021,Yu_2022_CVPR}. While most previous works are supervised and minimize the loss to a contact physiological measurement, we perform non-contrastive learning using only the model's estimated waveform.

The key realization is that we can place strong priors on the estimated pulse regarding its bandwidth and periodicity.
Observed signals outside the desired frequency range are pollutants, so penalizing the model for carrying them through the forward pass results in invariances to such noisy visual features.
We find that it is surprisingly easy to impose the desired constraints in the frequency domain.
Thus, all waveforms are transformed into their discrete Fourier components with the FFT before computing all losses in our approach.
Specifically, we calculate their power spectral density, $F = |\mathrm{FFT}(y)|^2$.
In our experiments, we set the input signal’s length to achieve a frequency resolution of $0.3\overline{3}$ bpm (\ie the \textit{n} or \textit{nfft} variable in some packages was set to 5,400).
The following sections describe the loss functions and augmentations used during training.

\subsection{Losses}

\begin{figure*}
    \centering
    \includegraphics[width=\linewidth]{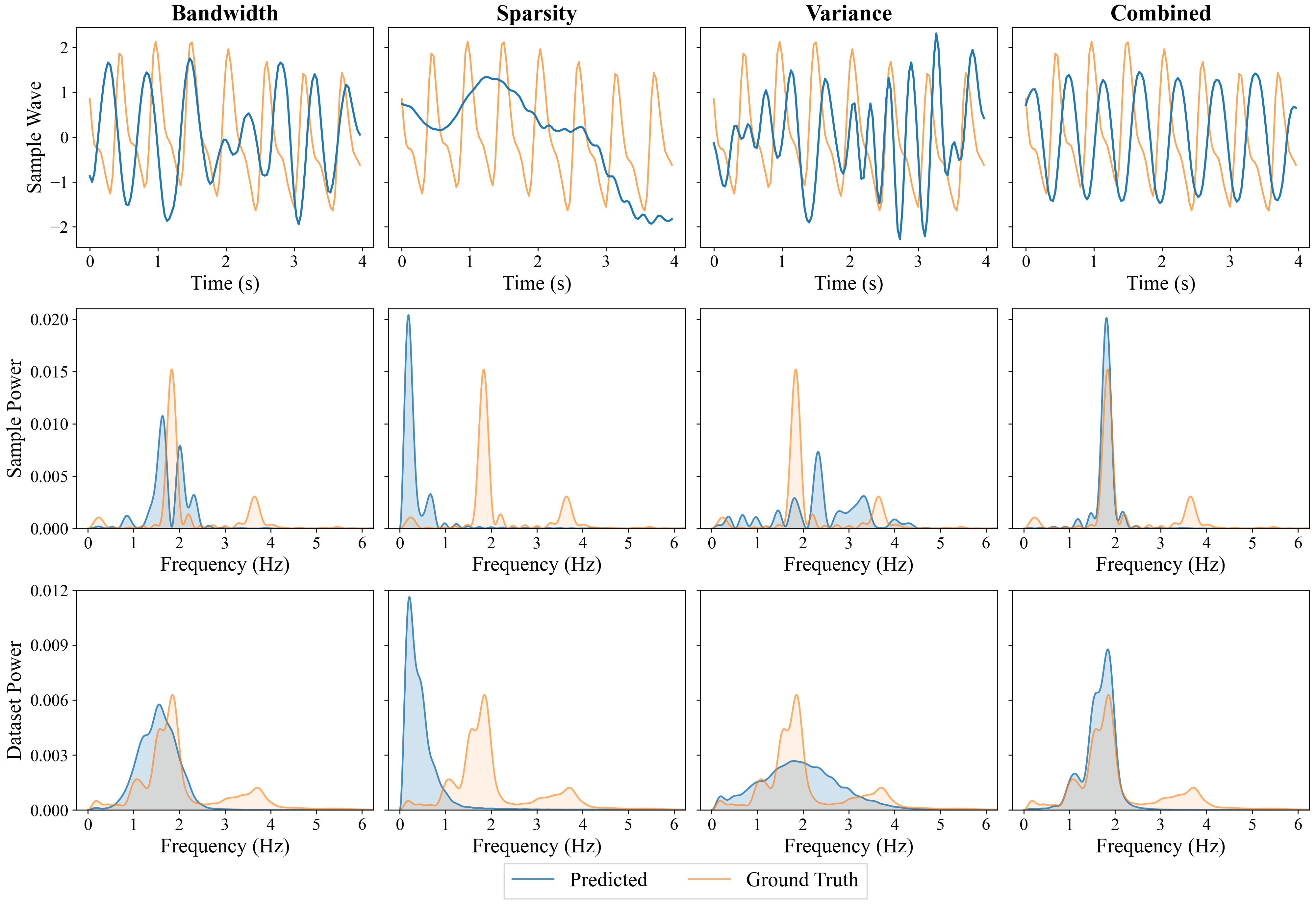}
    \caption{Each column shows predictions from models trained with one or all of the losses for 20 epochs on UBFC-rPPG. The first two rows show a sample in the time and frequency domain, respectively. The last row shows the signal power over the validation set computed by taking the sum of normalized power spectral densities from each sample, then dividing the result by the number of validation samples. The \textbf{bandwidth} loss penalizes signal power outside predefined bandlimits (40 to 180 bpm) to constrain the output space. The \textbf{sparsity} loss encourages a narrow spectrum containing strong periodicity. The \textbf{variance} loss encourages diverse power spectra in a batch, preventing the model from collapsing to a narrow bandwidth. When \textbf{combined}, the model estimates periodic signals within the desired bandlimits.}
    \label{fig:losses}
\end{figure*}

One of the advantages of unsupervised learning for periodic signals is that we can constrain the solution space significantly. For physiological signals such as respiration and blood volume pulse, we know the healthy upper and lower bounds of the frequencies. We also desire the extracted signal to be sparse in the frequency domain, and that our model filters out noise signals present in the video. With these constraints, we can greatly simplify the problem of finding good features for the desired signal in the data.

\textbf{Bandwidth Loss.} One of the most powerful constraints we can place on the model is frequency bandlimits. Past unsupervised methods have used the irrelevant power ratio (IPR) as a validation metric~\cite{Gideon_2021_ICCV,Gideon_2021_ICCVW,Sun_2022_ECCV} for model selection. We find that it is also effective during model training. The IPR penalizes the model for generating signals outside the desired bandlimits. With lower and upper bandlimits of $a$ and $b$, respectively, our bandwidth loss becomes:

\begin{equation}\label{eq:L_b}
    L_b = \frac{1}{\sum\limits_{i=-\infty}^\infty F_i}\Bigg[\sum\limits_{i=-\infty}^a F_i + \sum\limits_{i=b}^{\infty} F_i\Bigg],
\end{equation}

\noindent
where $F_i$ is the power in the $i$th frequency bin of the predicted signal. This simple loss enforces learning of many invariants, such as movement from respiration, talking, or facial expressions which typically occupy low frequencies. In our experiments we specify the limits as $a=0.\overline{66}$ Hz to $b=3$ Hz, which corresponds to a common pulse rate range from 40 bpm to 180 bpm. The first column of Fig. \ref{fig:losses} shows the result of training exclusively with the bandwidth loss $L_b$. The last row shows that the model concentrates signal power between the bandlimits.

\textbf{Sparsity Loss.} The pulse rate is the most common physiological marker associated with the blood volume pulse. Since we are primarily interested in the frequency, we can further improve our model by preventing wideband predictions. This also reveals the true signal we aim to discover by ignoring visual dynamics that are not strongly periodic.

We penalize energy within the bandlimits that is not near the spectral peak:

\begin{equation}\label{eq:L_s}
    L_s = \frac{1}{\sum\limits_{i=a}^b F_i}\Bigg[\sum\limits_{i=a}^{F^* - \Delta_F} F_i + \sum\limits_{i=F^* + \Delta_F}^b F_i\Bigg],
\end{equation}

\noindent
where $F^*=\mathrm{argmax}(F)$ and $\Delta_F$ are the frequencies of the spectral peak and padding around the peak, respectively.
For all rPPG experiments $\Delta_F=0.1$ Hz (or 6 beats per minute)~\cite{Nowara_BOE_2021}. Figure \ref{fig:losses} shows the result of training only with the sparsity loss in the second column. For the whole dataset, the power spectrum is sparsely distributed in the low frequencies, effectively filtering frequencies higher than 1 Hz. 

\textbf{Variance Loss.} One of the risks of non-contrastive methods is the model collapsing into trivial solutions and making predictions independently of the input features. In regularized methods such as VICReg~\cite{bardes2022vicreg}, a hinge loss on the variance over a batch of predictions is used to enforce diverse outputs. We use a similar strategy to avoid model collapse, but instead spread the variance in power spectral densities towards a uniform distribution over the desired frequency band.
Our variance loss processes a uniform prior distribution $P$ over $d$ frequencies, and a batch of $n$ spectral densities, $F = [v_1,...,v_n]$, where each vector is a $d$-dimensional frequency decomposition of a predicted waveform. We calculate the normalized sum of densities over the batch, $Q$, and define the variance loss as the squared Wasserstein distance~\cite{hou_emd_2017} to the uniform prior:

\begin{equation}
    L_v = \frac{1}{d} \sum_{i=1}^d \left( \mathrm{CDF}_i(Q) - \mathrm{CDF}_i(P) \right)^2,
\end{equation}

\noindent
where CDF is a cumulative distribution function. The third column of Fig. \ref{fig:losses} shows the effect of the variance loss during training. For a single sample, wide-band signals containing multiple frequencies are predicted, and the predicted frequencies cover the task's bandwidth. In our experiments we use a batch size of 20 samples. See the suppl. materials for an ablation experiment on the impact of batch size.

\textbf{Combining All Losses.}
Summarizing, our training loss function is a simple sum of the aforementioned losses:
\begin{equation}
    L = L_b + L_s + L_v.
\end{equation}
While one could weight particular components of the loss more than others, we specifically formulated the losses to scale them between 0 and 1. In our experiments, we find that a simple summation without weighting gives good performance. \textbf{The combined loss function encourages the model to search over the supported frequencies to discover visual features for a strong periodic signal.}
Remarkably, we find that this simple framework is sufficient for learning to regress subtle periodic signals such as the the blood volume pulse from video, as shown in the last column of Fig. \ref{fig:losses}.

\subsection{Augmentations}
Unlike Gideon \etal's~\cite{Gideon_2021_ICCV} approach, which only applies frequency augmentations, we apply several augmentations to both the spatial and temporal dimensions to learn invariances to noisy visual signals. In fact, we found that without augmentations, models did not converge during training (see the supplementary material).

\textbf{Image Intensity Augmentations.} Gaussian noise is added to each pixel with zero mean and a standard deviation of 2 on an image scale from 0 to 255. Each clip is darkened or brightened by adding a constant from a Gaussian distribution with mean 0 and standard deviation of 10.

\textbf{Spatial Augmentations.} We randomly horizontally flip a video clip with 50\% probability. The spatial dimension of a clip are randomly square cropped down to between half the original length and the original length. The cropped clip is then linearly interpolated back to the original dimensions.

\textbf{Temporal Augmentations.} With the general assumption that the desired signal is strongly periodic and sparsely represented in the Fourier domain, we randomly flip a video clip along the time dimension with a probability of 50\%. Note that the Fourier decomposition of a time-reversed sinusoid is identical to that of the original sinusoid.

\textbf{Frequency Augmentations.} Perhaps the most important augmentation is frequency resampling~\cite{Gideon_2021_ICCV}, where the video is linearly interpolated to a different frame rate. This augmentation is particularly interesting for rPPG, because it transforms the video input and target signal equivalently along the time dimension, making it equivariant. Given the aforementioned transformations that are invariant, $\tau(\cdot) \sim \mathcal{T}$, the equivariant frequency resampling operation, $\phi(\cdot) \sim \Phi$, and a model $f(\cdot)$ that infers a waveform from a video we have the following:

\begin{equation}
    \phi(f(\tau(x))) = f(\phi(\tau(x))).
\end{equation}

This is a powerful augmentation, because it allows us to augment the target distribution along with the video input. In our experiments we randomly resample input clips by a factor $c \sim U(0.6, 1.4)$. After applying the resampling augmentation, we scale the bandlimits by $c$, to avoid penalizing the model if the augmentation pushed the underlying pulse frequency outside of the original bandlimits.


\section{Datasets}

\begin{figure}
    \centering
    \includegraphics[width=\linewidth]{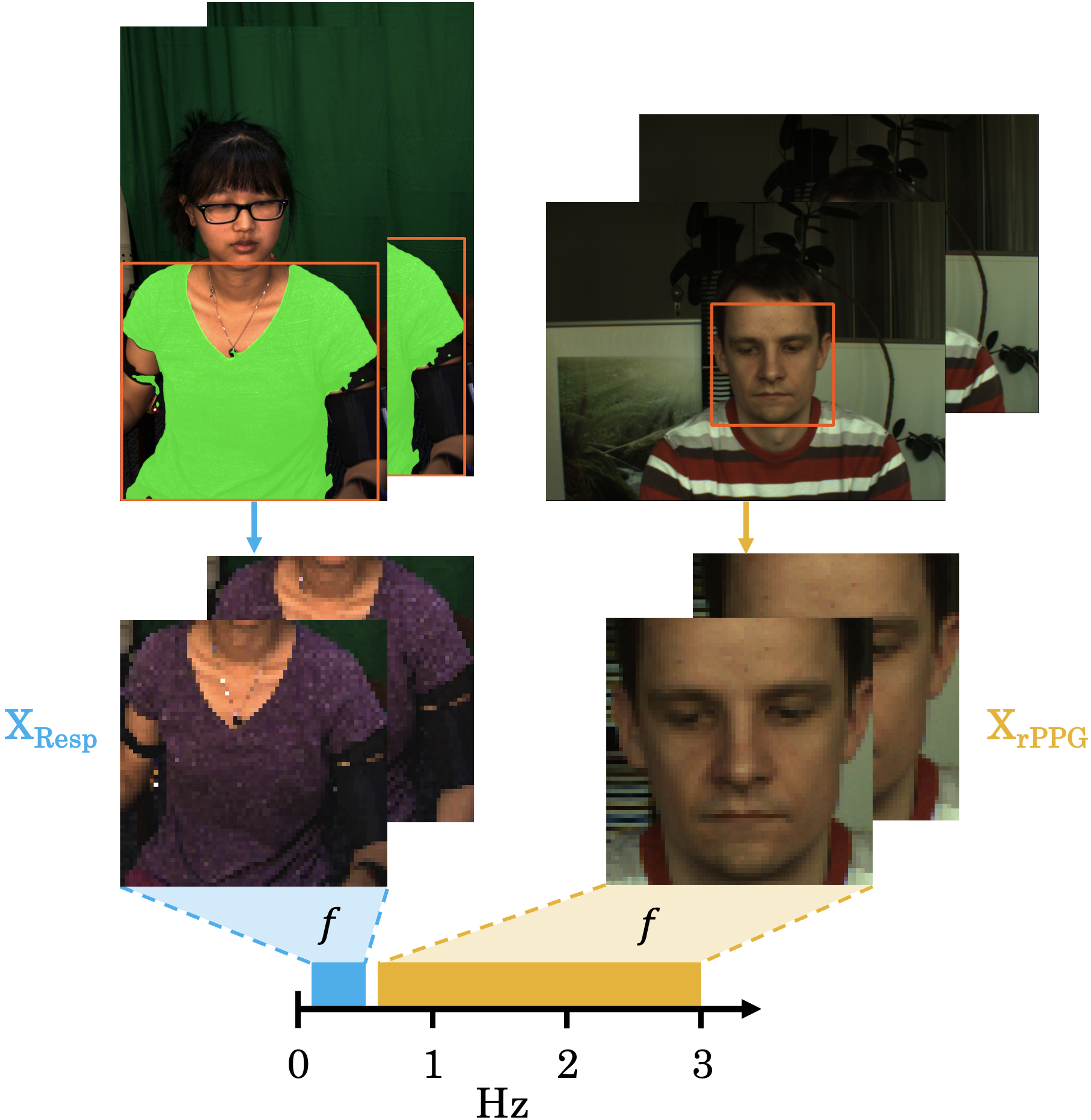}
    \caption{Preprocessing steps for remote respiration (left) and pulse estimation (right), along with the bandlimits used during training with SiNC.}
    \label{fig:preprocessing}
\end{figure}

We use PURE~\cite{Stricker2014}, UBFC-rPPG~\cite{Bobbia2019}, and DDPM~\cite{Speth_IJCB_2021} as benchmark rPPG datasets for training and testing, and CelebV-HQ dataset~\cite{zhu2022celebvhq} and HKBU-MARs~\cite{Liu_2016_CVPRW} for unsupervised training only.
For remote respiration experiments we use the MSPM~\cite{Speth_TIM_2023} dataset.

\textbf{Deception Detection and Physiological Monitoring (DDPM)}~\cite{Speth_IJCB_2021,Vance2022} consists of data from 86 subjects attempting to answer questions deceptively. Interviews were recorded at 90 frames-per-second for more than 10 minutes on average. Natural conversation and head pose changes make it a difficult and less-constrained rPPG dataset.

\textbf{PURE}~\cite{Stricker2014} is a benchmark rPPG dataset consisting of 10 subjects recorded over 6 sessions. Each session lasted approximately 1 minute, and raw video was recorded at 30 fps. The 6 sessions for each subject consisted of: (1) steady, (2) talking, (3) slow head translation, (4) fast head translation, (5) small and (6) medium head rotations. Pulse rates are at or close to the subject's resting rate.

\textbf{UBFC-rPPG}~\cite{Bobbia2019} contains 1-minute long videos from 42 subjects recorded at 30 fps. Subjects played a time-sensitive mathematical game to raise their heart rates, but head motion is limited during the recording.

\textbf{HKBU 3D Mask Attack with Real World Variations (HKBU-MARs)}~\cite{Liu_2016_CVPRW} consists of 12 subjects captured over 6 different lighting configurations with 7 different cameras each, resulting in 504 videos lasting 10 seconds each. The diverse lighting and camera sensors make it a valuable dataset for unsupervised training. We use version 2 of HKBU-MARs, which contains videos with both realistic 3D masks and unmasked subjects.

\textbf{High-Quality Celebrity Video Dataset (CelebV-HQ)}~\cite{zhu2022celebvhq} is a set of processed YouTube videos containing 35,666 face videos from over 15,000 identities. The videos vary dramatically in length, lighting, emotion, motion, skin tones, and camera sensors. The greatest challenge in harnessing online videos is their reduced quality due to compression before upload and by the video provider. Compression is a known challenge for rPPG, since the blood volume pulse is so subtle optically~\cite{McDuff2017,Nowara_ICCVW_2019,Rapczynski2019,Nowara_BOE_2021}.

\textbf{Multi-Site Physiological Monitoring (MSPM)}~\cite{Speth_TIM_2023} is a large video dataset consisting of 103 subjects with ground truth pulse, blood pressure, and respiration.
We used this dataset for our remote respiration experiments.
The respiration ground truth was collected by having subjects follow a 120 second long video with cues on the screen to inhale and exhale.
The breathing frequencies were modulated between 0.167-0.333 Hz (10-20 breaths per minute), which is considered a healthy range for adults~\cite{russo2017physiological}.
We used the videos from the ``RGB Front" camera. The entire video including activities other than the respiration activity was used for training.
\section{Training Details}

\subsection{rPPG Data Preprocessing}
To prepare the video clips for the spatiotemporal deep learning models, we first extract 68 face landmarks with OpenFace~\cite{Baltrusaitis2018}. We then define a bounding box in each frame with the minimum and maximum $(x,y)$ locations by extending the crop horizontally by 5\% to ensure that the cheeks and jaw are present. The top and bottom are extended by 30\%  and 5\% of the bounding box height, respectively, to include the forehead and jaw. We further extend the shorter of the two axes to the length of the other to form a square. The cropped frames are then resized to 64$\times$64 pixels with bicubic interpolation.
An example of the preprocessing for rPPG is shown on the right side of Fig.~\ref{fig:preprocessing}.
For faster processing of the massive CelebV-HQ~\cite{zhu2022celebvhq} dataset, we instead use MediaPipe Face Mesh~\cite{lugaresi2019mediapipe} for landmarking.
For rPPG, each input sample is $T=120$ frames (4 seconds) in duration.

\subsection{Model Architectures}
For most of our experiments we use a 3D-CNN architecture similar to \cite{Speth_CVIU_2021} without temporal dilations, which was derived from PhysNet~\cite{Yu2019}.
We use a temporal kernel width of 5, and replace default zero-padding by repeating the edges.
Zero-padding along the time dimension can result in edge effects that add artificial frequencies to the predictions.
Early experiments showed that temporal dilations caused aliasing and reduced the bandwidth of the model to specific frequencies.

Our losses and framework may theoretically be applied to any task and architecture with dense predictions along one or more dimensions.
However, popular rPPG architectures such as DeepPhys~\cite{Chen2018} may be ill-suited for the approach, since they only consume frame differences, and the number of time points should be large enough to give sufficient frequency resolution with the FFT.
To show that our approach can generalize to different architectures, we ran additional experiments with the temporal shift convolutional attention network (TS-CAN) architecture~\cite{Liu_MTTS_2020}.
TS-CAN is a two-stream network that takes RGB frames in one stream to compute attention masks (visual branch) and frame differences in the other where the motion is applied (motion branch).
The convolutional operations in TS-CAN are 2-dimensional, making it a relatively lightweight model.

\subsection{Supervised Training}
To properly compare our approach to its supervised counterpart we use the same model architecture and train it with the commonly used negative Pearson loss between the predicted waveform and the contact sensor ground truth~\cite{Yu2019}.
During training we apply all of the same augmentations except time reversal.
When training the TS-CAN models, we only use flipping, illumination, and random cropping, as we found that the model could not converge with Gaussian noise or frequency augmentations. 
We use the AdamW~\cite{Loshchilov2019AdamW} optimizer with a learning rate of 0.0001 for both supervised and unsupervised training.
Models are trained for 200 epochs on PURE and UBFC-rPPG, and for 40 epochs on DDPM.
The model from the epoch with the lowest loss on the validation set is selected for testing.

\subsection{Unsupervised Training}
Unsupervised models are trained for the same number of epochs as the supervised setting for both PURE and UBFC-rPPG, but we train for an additional 40 epochs on DDPM, since this dataset is considerably more difficult. We set the batch size to 20 samples during training. Contrary to previous unsupervised approaches~\cite{Gideon_2021_ICCV,Sun_2022_ECCV}, we leverage validation sets for model selection by selecting the model with the lowest combined bandwidth and sparsity losses. The creation of the dataset splits is described in the next section.

\subsection{Evaluation}
Pulse rates are computed as the highest spectral peak between 0.66 Hz and 3 Hz (equivalent to 40 bpm to 180 bpm) over a 10-second sliding window. The same procedure is applied to the ground truth waveforms for a reliable evaluation~\cite{Mironenko2020}.
Respiration rates are computed as the spectral peak between $0.16\Bar{6}$ Hz and $0.3\Bar{3}$ Hz (10-20 breaths per minute) over a 30-second sliding window.
We apply common error metrics such as mean absolute error (MAE), root mean square error (RMSE), and Pearson correlation coefficient ($r$).

We perform 5-fold cross validation for both PURE and UBFC with the same folds as \cite{Gideon_2021_ICCV}, and use the predefined dataset splits from DDPM~\cite{Speth_CVIU_2021}. Differently from \cite{Gideon_2021_ICCV}, we use 3 of the folds for training, 1 for validation, and the remaining for testing rather than only training and testing partitions. We train 3 models with different initializations, resulting in 15 models trained on PURE and UBFC each, and 3 models trained on DDPM. We present the mean and standard deviation of the errors in the results.

\section{Results}
\label{sec:results}
\setlength\tabcolsep{4pt}
\begin{table*}[htb!]
\caption{\textbf{Intra-dataset} pulse rate estimation results. The best results for a given metric/dataset are bolded, and the second-best results are underlined. For a better comparison with results in the literature, this table follows \cite{Sun_2022_ECCV}. MAE: Mean Absolute Error; RMSE: Root Mean Square Error; $r$: Pearson correlation coefficient.}
\fontsize{7.533}{9}\selectfont
\centering
{
\begin{threeparttable}
\begin{tabular}{llccccccccc}
\toprule
\multirow{2}{*}{\begin{tabular}[c]{@{}l@{}}\\ Types\end{tabular}} &
\multirow{2}{*}{\begin{tabular}[c]{@{}l@{}}\\Methods\end{tabular}} &
\multicolumn{3}{c}{UBFC-rPPG} & \multicolumn{3}{c}{PURE} &
\multicolumn{3}{c}{DDPM} \\ 
\cmidrule(lr){3-5}\cmidrule(lr){6-8}\cmidrule(lr){9-11}
& & \begin{tabular}[c]{@{}c@{}}MAE\\ (bpm)\end{tabular} & \begin{tabular}[c]{@{}c@{}}RMSE\\ (bpm)\end{tabular} & $r$                    & \begin{tabular}[c]{@{}c@{}}MAE\\ (bpm)\end{tabular} & \begin{tabular}[c]{@{}c@{}}RMSE\\ (bpm)\end{tabular} & $r$                    & \begin{tabular}[c]{@{}c@{}}MAE\\ (bpm)\end{tabular} & \begin{tabular}[c]{@{}c@{}}RMSE\\ (bpm)\end{tabular} & $r$                  \\ 
\midrule
\multirow{5}{*}{\rot{Traditional}}
& GREEN \cite{Verkruysse2008} & 7.50 & 14.41 & 0.62 & 7.23 & 17.05 & 0.69 & 32.79 & 43.09 & 0.04 \\
& ICA \cite{Poh2011} & 5.17 & 11.76 & 0.65 & 3.76 & 12.60 & 0.85 & 22.22 & 35.77 & 0.40 \\
& CHROM \cite{DeHaan2013} &
    2.36 & 9.23 & 0.87 &
    0.75 & 2.23 & \bf 1.00 &
    13.48 & 28.53 & 0.56 \\
& 2SR \cite{Wang2016} & 6.90 & 18.50 & 0.65 &
    2.44 & 3.06 & 0.98 & 22.08 & 39.73 & 0.22 \\
& POS \cite{Wang2017} &
    2.11 & 9.11 & 0.87 &
    0.80 & 4.11 & 0.98 &
    9.03 & 23.07 & 0.70 \\
 
\midrule
\multirow{7}{*}{\rot{Supervised}}

& HR-CNN \cite{Spetlik_2018} & - & - & - & 1.84 & 2.37 & 0.98 & - & - & - \\
& SynRhythm \cite{Niu2018} & 5.59 & 6.82 & 0.72 & - & - & - & - & - & - \\
& PulseGAN \cite{Song_2021} & 1.19 & 2.10 & \underline{0.98} & - & - & - & - & - & - \\
& Dual-GAN \cite{Lu_2021} &
    \underline{0.44} & \underline{0.67} & \bf 0.99 &
    0.82 & \underline{1.31} & \underline{0.99} &
    - & - & - \\
& RPNet \cite{Speth_CVIU_2021}\tnote{\textdagger} &
    0.53 $\pm$ 0.01 & 1.78 $\pm$ 0.02 & \bf{0.99 $\pm$ 0.00} &
    1.15 $\pm$ 0.27 & 5.77 $\pm$ 1.25 & 0.96 $\pm$ 0.02 &
    \bf 3.46 $\pm$ 0.24 & \bf 12.47 $\pm$ 0.68 & \bf 0.91 $\pm$ 0.01 \\
& PhysNet \cite{Yu2019}\tnote{\textdagger} &
    0.55 $\pm$ 0.03 & 2.03 $\pm$ 0.37 & \bf 0.99 $\pm$ 0.00 &
    0.99 $\pm$ 0.19 & 5.22 $\pm$ 0.93 & 0.97 $\pm$ 0.01 &
    \underline{3.96 $\pm$ 0.76} & \underline{13.57 $\pm$ 1.74} & 
    \underline{0.89 $\pm$ 0.03} \\
& TSCAN \cite{Liu_MTTS_2020}\tnote{\textdagger} &
    1.42 $\pm$ 0.39 & 7.43 $\pm$ 2.52 & 0.90 $\pm$ 0.05 &
    4.44 $\pm$ 0.26 & 19.13 $\pm$ 0.66 & 0.72 $\pm$ 0.01 &
    7.66 $\pm$ 0.14 & 21.16 $\pm$ 0.38 & 0.76 $\pm$ 0.01 \\
\midrule

\multirow{3}{*}{\rot{\begin{tabular}[c]{@{}c@{}}Self-\\Super-\\vised\tnote{*}\end{tabular}}}
 & SLF-RPM~\cite{Wang_SSL_2022} &
    8.39 & 9.70 & 0.70 &
    - & - & - &
    - & - & - \\
 & SimPer~\cite{Yuzhe_SimPer_2022} &
    4.24 & - & - &
    3.89 & - & - &
    - & - & - \\
 & rPPG-MAE~\cite{Liu2023rPPGMAE} &
    \bf 0.17 & \bf 0.21 & \bf 0.99 &
    \bf 0.40 & \bf 0.92 & \underline{0.99} &
    - & - & - \\
\midrule

\multirow{5}{*}{\rot{\begin{tabular}[c]{@{}c@{}}Unsuper-\\vised\end{tabular}}}
 & Gideon2021 \cite{Gideon_2021_ICCV} & 
    1.85 & 4.28 & 0.93 &
    2.3 & 2.9 & \underline{0.99} &
    - & - & - \\
 & Contrast-Phys~\cite{Sun_2022_ECCV} &
    0.64 & 1.00 & \bf 0.99 &
    1.00 & 1.40 & \underline{0.99} &
    9.70 $\pm$ 2.90 & 25.02 $\pm$ 6.01 & 0.58 $\pm$ 0.19 \\
 & Yue2023~\cite{Yue2023} &
    0.58 & 0.94 & \bf 0.99 &
    1.23 & 2.01 & \underline{0.99} &
    - & - & - \\
 & \bf{\ourapproach (PhysNet)} &
    0.59 $\pm$ 0.00 & 1.83 $\pm$ 0.04 & \bf 0.99 $\pm$ 0.00 &
    \underline{0.61 $\pm$ 0.06} & 1.84 $\pm$ 0.40 & \bf 1.00 $\pm$ 0.00 &
    5.87 $\pm$ 0.11 & 17.44 $\pm$ 0.16 & 0.81 $\pm$ 0.00 \\
 & \bf{\ourapproach (TSCAN)} &
    0.84 $\pm$ 0.01 & 3.72 $\pm$ 0.14 & 0.97 $\pm$ 0.00 &
    1.11 $\pm$ 0.05 & 5.59 $\pm$ 0.31 & 0.97 $\pm$ 0.00 &
    4.09 $\pm$ 0.30 & 13.67 $\pm$ 0.68 & \underline{0.89 $\pm$ 0.01} \\
\bottomrule
\end{tabular}
\begin{tablenotes}\footnotesize
        \item[*] Self-supervised methods require fine-tuning on labeled data with a linear classifier after pretraining.
        \item[\textdagger] Some supervised methods were trained with identical data augmentations to \ourapproach for fair comparison.
        \end{tablenotes}
    \end{threeparttable}
}
\label{tab:within_dataset}
\end{table*}

\subsection{Within-Dataset Testing}
Table \ref{tab:within_dataset} shows the results for models trained and tested on subject-disjoint partitions from the same datasets. For PURE and UBFC we achieve MAE lower than 1 bpm, performing better or on par with all traditional and supervised learning approaches. For PURE, our approach gives the lowest MAE and a Pearson $r$ of nearly 1. Performance drops on DDPM due to the overall difficulty of the dataset. SiNC outperforms contrastive approaches, only being surpassed by supervised deep learning models.

In comparison to other unsupervised methods, Contrast-Phys~\cite{Sun_2022_ECCV} gives the most competitive performance on all but DDPM. Note that our approach gives the lowest MAE on all datasets, but has higher RMSE. We believe this is due their use of harmonic removal as a post-processing step when estimating the pulse rate, which is not described in \cite{Sun_2022_ECCV}, but can be found in their publicly available code. 

\subsection{Cross-Dataset Testing}
\label{sec:cross_dataset}
\begin{table}[!htb]
\setlength\tabcolsep{5pt}
\caption{\textbf{Cross-dataset} pulse rate estimation performance. The top 3 training datasets are common rPPG benchmarks, while HKBU was not designed for rPPG and has no pulse ground truth. Note that testing on UBFC after training on DDPM performs well, since their frequency distributions are similar, while PURE's pulse rates tend to be much lower.}
\fontsize{7.533}{9}\selectfont
\begin{tabular}{cclrr}
\toprule
\begin{tabular}{@{}c@{}}\textbf{Training}\\\textbf{Dataset}\end{tabular} & \begin{tabular}{@{}c@{}}\textbf{Testing}\\\textbf{Dataset}\end{tabular} & {\bf Method} & \begin{tabular}{@{}c@{}}{\bf MAE\phantom{xxi}}\\\textbf{(bpm)\phantom{xxi}}\end{tabular} & {\bf $r$\phantom{xxxx}}
\\
\cmidrule{1-5}
\multirow{8}{*}{DDPM}
& UBFC & SiNC+TTA (PhysNet) & \bf 0.54 $\pm$ 0.02 & \bf 1.00 $\pm$ 0.00\\
& UBFC & \ourapproach (PhysNet) & 0.88 $\pm$ 0.25 & 0.98 $\pm$ 0.01\\
& UBFC & \ourapproach (TSCAN) & 1.20 $\pm$ 0.18 & 0.94 $\pm$ 0.01\\
& UBFC & Contrast-Phys & 1.14 $\pm$ 0.38 & 0.96 $\pm$ 0.03\\
& UBFC & PhysNet & 1.11 $\pm$ 0.42 & 0.95 $\pm$ 0.02 \\
\cline{2-5}\\[-2ex]
& PURE & SiNC+TTA (PhysNet) & \bf 0.54 $\pm$ 0.04 & \bf 1.00 $\pm$ 0.00\\
& PURE & \ourapproach (PhysNet) & 3.12 $\pm$ 1.07 & 0.88 $\pm$ 0.05 \\
& PURE & \ourapproach (TSCAN) & 0.77 $\pm$ 0.02 & 0.98 $\pm$ 0.00 \\
& PURE & Contrast-Phys & 13.02 $\pm$ 6.12 & 0.19 $\pm$ 0.59  \\
& PURE & PhysNet & 1.46 $\pm$ 0.34 & 0.95 $\pm$ 0.02 \\
\midrule
 
\multirow{8}{*}{UBFC}
& DDPM & SiNC+TTA (PhysNet) & \bf 5.10 $\pm$ 0.37 & \bf 0.88 $\pm$ 0.01 \\
& DDPM & \ourapproach (PhysNet) & 18.53 $\pm$ 0.36 & 0.38 $\pm$ 0.01 \\
& DDPM & \ourapproach (TSCAN) & 19.51 $\pm$ 0.33 & 0.41 $\pm$ 0.01 \\
& DDPM & Contrast-Phys & 22.93 $\pm$ 1.02 & 0.18 $\pm$ 0.04 \\
& DDPM & PhysNet & 18.58 $\pm$ 0.12 & 0.40 $\pm$ 0.00 \\
\cline{2-5}\\[-2ex]
& PURE & SiNC+TTA (PhysNet) & \bf 1.02 $\pm$ 0.19 & \bf 0.98 $\pm$ 0.01 \\
& PURE & \ourapproach (PhysNet) & 4.02 $\pm$ 0.06 & 0.86 $\pm$ 0.00 \\
& PURE & \ourapproach (TSCAN) & 1.15 $\pm$ 0.03 & 0.96 $\pm$ 0.00 \\
& PURE & Contrast-Phys & 19.61 $\pm$ 2.01 & 0.33 $\pm$ 0.06 \\
& PURE & PhysNet & 3.81 $\pm$ 0.34 & 0.87 $\pm$ 0.02 \\
\midrule

\multirow{8}{*}{PURE}
& UBFC & SiNC+TTA (PhysNet) & \bf 0.98 $\pm$ 0.36 & \bf 0.95 $\pm$ 0.04 \\
& UBFC & \ourapproach (PhysNet) & 6.64 $\pm$ 1.76 & 0.59 $\pm$ 0.10 \\
& UBFC & \ourapproach (TSCAN) & 1.42 $\pm$ 0.03 & 0.94 $\pm$ 0.00 \\
& UBFC & Contrast-Phys & 10.22 $\pm$ 0.38 & 0.45 $\pm$ 0.04 \\
& UBFC & PhysNet & 7.02 $\pm$ 3.35 & 0.60 $\pm$ 0.13 \\
\cline{2-5}\\[-2ex]
& DDPM & SiNC+TTA (PhysNet) & \bf 5.31 $\pm$ 1.11 & \bf 0.86 $\pm$ 0.04 \\
& DDPM & \ourapproach (PhysNet) & 24.92 $\pm$ 0.65 & 0.20 $\pm$ 0.00 \\
& DDPM & \ourapproach (TSCAN) & 22.87 $\pm$ 0.15 & 0.31 $\pm$ 0.00 \\
& DDPM & Contrast-Phys & 29.63 $\pm$ 0.48 & 0.03 $\pm$ 0.02 \\
& DDPM & PhysNet & 28.03 $\pm$ 2.20 & 0.13 $\pm$ 0.05\\
\midrule

\multirow{3}{*}{\begin{tabular}{@{}c@{}}HKBU\\(non-\\rPPG)\end{tabular}}

& UBFC & \ourapproach (PhysNet) &  1.08 $\pm$ 0.03 & 0.95 $\pm$ 0.00 \\
& PURE      & \ourapproach (PhysNet) &  2.43 $\pm$ 0.20 & 0.90 $\pm$ 0.02 \\
& DDPM      & \ourapproach (PhysNet) & 20.34 $\pm$ 0.25 & 0.19 $\pm$ 0.02 \\
\bottomrule
\end{tabular}
\label{tab:cross_dataset}
\end{table}

We perform cross-dataset testing to analyze whether the approach is robust to changes to the lighting, camera sensor, pulse rate distribution, and motion. Table \ref{tab:cross_dataset} shows the results for SiNC and supervised training with the same architecture. We find that the performance is similar for the supervised and unsupervised approaches when transferring to different data sources. Training on PURE exclusively gives relatively poor results when transferring to UBFC-rPPG and DDPM, due to the low pulse rate variability within PURE samples and lack of movement. Training on DDPM gives the best results overall, since the dataset is the largest and captures larger subjects' movements compared to other datasets.

\subsection{Training with non-rPPG Datasets}

Given the abundance of face videos publicly available online, we trained a model on the \textbf{CelebV-HQ} dataset~\cite{zhu2022celebvhq}. After processing the available videos with MediaPipe and resampling to 30 fps, our unlabeled dataset consisted of 34,029 videos. We trained the model for 23 epochs and manually stopped training due to a plateau in the validation loss.
Unfortunately, we found that the model could not converge to the true blood volume pulse. We attribute the failure to poor video quality from compression. Although the videos were downloaded with the highest available quality, they have likely been compressed, removing the pulse signal entirely. See the supplementary material for an assessment of rPPG quality using POS~\cite{Wang2017} on CelebV-HQ.

The \textbf{HKBU-MARs} dataset~\cite{Liu_2016_CVPRW} was designed for face presentation attack detection, but we trained models on the ``real'' video sessions in the dataset. The bottom rows in Table \ref{tab:cross_dataset} show the results for training on HKBU-MARs, then testing on the benchmark rPPG datasets. Training on HKBU-MARs gives better results when testing on UBFC-rPPG and PURE than all training sets except DDPM, which is an order of magnitude larger. \textbf{To our knowledge, this is the first succesful experiment showing that non-rPPG videos can be used to train robust rPPG models, even if they do not have ground-truth pulse labels}.

\begin{figure}
    \centering
    \includegraphics[width=\linewidth]{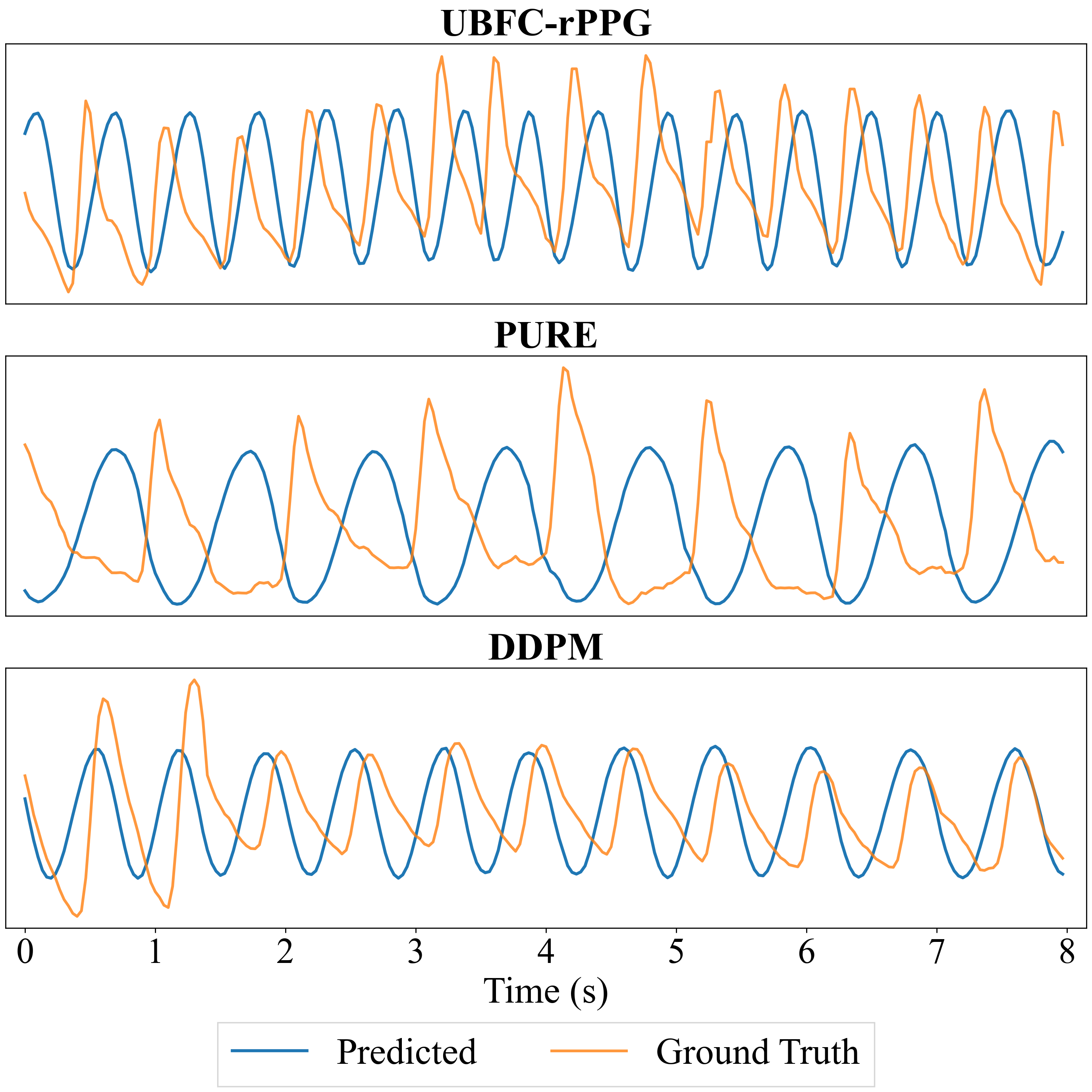}
    \caption{Within-dataset waveform predictions on all baseline datasets from end-to-end unsupervised models over an 8-second window. The model predictions are remarkably periodic without any form of filtering. Note that phase is not considered during training, so each model learns its own phase shift.}
    \label{fig:preds}
\end{figure}

\subsection{Ablation Study on Losses}
\begin{table}[!htb]
\setlength\tabcolsep{6pt}
\centering\footnotesize
\caption{Ablation study on the loss functions used during training. Results are shown for models trained and tested on UBFC-rPPG.}
\begin{tabular}{lrrr}
\toprule
{\bf Loss} & {\bf MAE (bpm)} & {\bf RMSE (bpm)} & {\bf $r$\phantom{xxxx}}
\\
\midrule
    $L_b$   &  3.08 $\pm$ 1.69 &  8.08 $\pm$ 3.61 &  0.87 $\pm 0.08$ \\
    $L_s$   & 45.50 $\pm$ 1.22 & 50.04 $\pm$ 0.94 & -0.04 $\pm 0.08$ \\
    $L_v$   & 22.89 $\pm$ 2.83 & 31.51 $\pm$ 2.36 &  0.22 $\pm 0.09$ \\
    $L_s+L_v$   & 51.24 $\pm$ 5.36 & 57.80 $\pm$ 7.39 & -0.04 $\pm 0.09$ \\
    $L_b+L_s$   &  9.99 $\pm$ 2.55 & 17.14 $\pm$ 2.36 &  0.51 $\pm 0.14$ \\
    $L_b+L_v$   &  4.18 $\pm$ 2.88 &  8.90 $\pm$ 5.24 &  0.82 $\pm 0.14$ \\
    \midrule
    \textbf{$L_b+L_s+L_v$}  & \bf 0.59 $\pm$ 0.00 & \bf 1.83 $\pm$ 0.04 & \bf 0.99 $\pm$ 0.00 \\
\bottomrule
\end{tabular}
\label{tab:loss_ablation}
\end{table}
We trained models using all combinations of loss components to analyze their contributions. Table \ref{tab:loss_ablation} shows the results for training and testing on UBFC-rPPG. The bandwidth loss is the most critical for discovering the true blood volume pulse, while the sparsity and variance losses do not learn the desired signal by themselves. Surprisingly, combining the bandwidth loss with just one of the sparsity or variance losses gives worse performance than just the bandwidth loss. However, when combining all three components, the model achieves impressive results.


\section{Personalization From Pretrained Models}
\label{sec:personalization}

\begin{table}\footnotesize
    \centering
    \caption{rPPG results after personalizing PhysNet models originally trained on the UBFC dataset to the first 20 seconds of each subject in the DDPM dataset.}
    \setlength{\tabcolsep}{3.0pt}
    \resizebox{\columnwidth}{!}{%
    \begin{tabular}{cccrr}
        \toprule
        Personalized & Dataset & Subset & MAE\phantom{xxii} & $r$\phantom{xxxxi} \\
        \midrule
        \xmark & UBFC & All 
        & 0.59 $\pm$ 0.00
        & 0.99 $\pm$ 0.00 \\
        \xmark & PURE & All
        & 4.02 $\pm$ 0.06
        & 0.86 $\pm$ 0.00 \\
        \xmark & DDPM & All
        & 18.53 $\pm$ 0.36
        & 0.38 $\pm$ 0.01 \\
        \midrule
        \cmark & DDPM & \textbf{Same Subject}
        & \textbf{6.36 $\pm$ 0.69}
        & \textbf{0.84 $\pm$ 0.02} \\
        \cmark & DDPM & Cross-Subject
        & 14.32 $\pm$ 0.64 
        & 0.51 $\pm$ 0.03 \\
        \cmark & DDPM & All
        & 13.60 $\pm$ 0.63
        & 0.54 $\pm$ 0.03 \\
        \cmark & UBFC & All
        & 1.51 $\pm$ 0.79
        & 0.94 $\pm$ 0.05 \\
        \cmark & PURE & All
        & 5.84 $\pm$ 0.84
        & 0.75 $\pm$ 0.03 \\
        \bottomrule
    \end{tabular}}
    \label{tab:finetuning_personalization}
\end{table}

This section describes the scenario where a pretrained model is finetuned on a small amount of unlabelled video from a single subject.
In effect, we perform model personalization at the very beginning of the video, then freeze the model for the remainder of inference on that subject.
Figure~\ref{fig:TTA_and_Personalization} shows an overview of model personalization with SiNC.
This experimental setup is useful in applications where the model cannot generalize well to the unseen subject, camera, lighting, or behavior in the new video during testing.
By finetuning the model on the short video sequence of the subject, we can expect the model to calibrate to the new environmental settings.
If the subject continued to use the system, this personalized model may outperform a single all-purpose model.

We perform experiments on one of the most challenging cross-dataset scenarios observed in Section~\ref{sec:cross_dataset}: training on UBFC and testing on DDPM.
For this scenario, both supervised and unsupervised approaches gave MAE values over 18 bpm, leaving ample room for improvement.
Specifically, we use the first 20 seconds of each test subject in DDPM as a training set for finetuning a single model.
We use the PhysNet models trained with SiNC on UBFC from our k-folds experiments as the initial weights.
Each model is trained for 50 epochs with a batch size of 20 samples using all but the Gaussian noise and frequency augmentations.
Here, an epoch is the number of 120 frame samples that can be made from the 20-second video with a 60-frame overlap.
After training, we had 15 personalized models for each subject in the DDPM test set, since we used 3 initializations over each of the 5 folds in our UBFC experiments.

Table~\ref{tab:finetuning_personalization} shows the results of training the personalized models.
Results on UBFC and PURE show that the model does undergo minor ``forgetting", where the performance on the original UBFC training dataset and PURE degrades.
However, personalized models improve drastically when tested on their corresponding subject.
The MAE drops by 12.17 bpm down to 6.36 bpm (65.7\% reduction) and the Pearson's $r$ correlation jumps from 0.38 to 0.84. In fact, the correlation for the personalized models is higher than the PhysNet models trained with SiNC on the entire DDPM training set.
\section{Test-Time Adaptation}
\label{sec:adaptation}

\begin{figure}
    \centering
    \includegraphics[width=\linewidth]{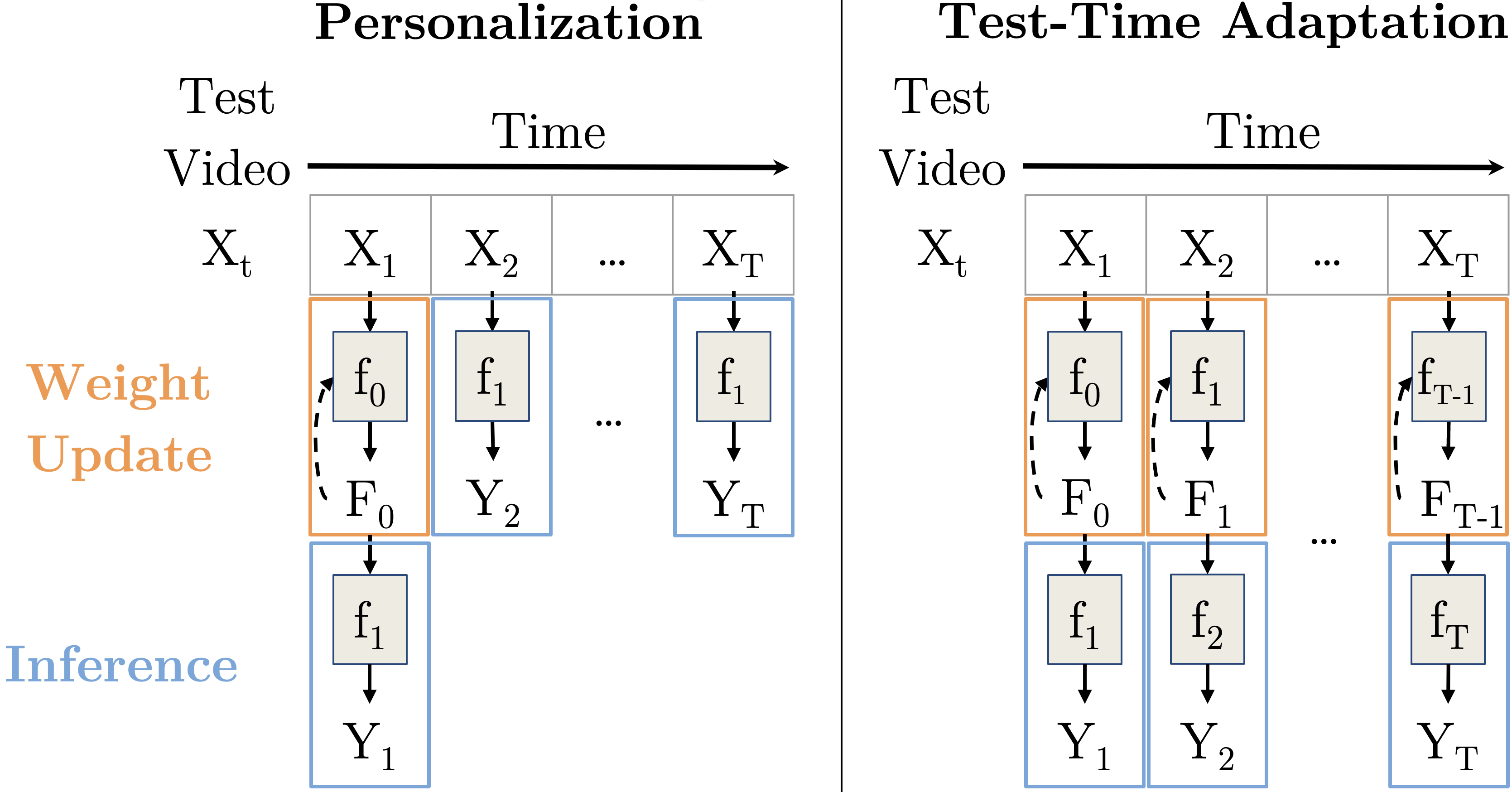}
    \caption{Overview of model personalization and test-time adaptation. We show the model, $f_t$ at each timestep $t$ along with the corresponding prediction, $Y_t$. The initial model, $f_0$, is pretrained with the SiNC framework in our experiments. Model weights are updated at various timesteps by minimizing the SiNC loss over the frequency prediction on that test sample, $L(F_t)$.
    }
    \label{fig:TTA_and_Personalization}
\end{figure}

\begin{figure}
    \centering
    \includegraphics[width=\linewidth]{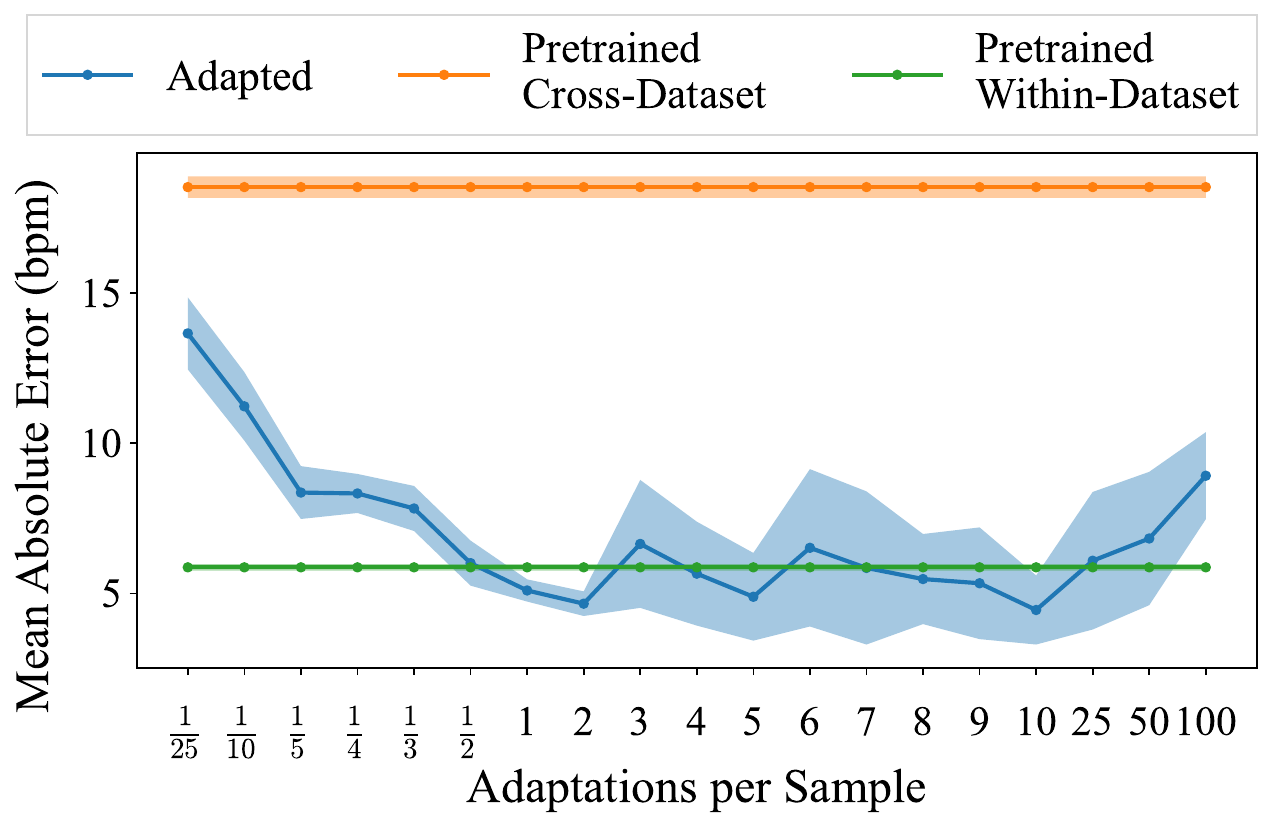}
    \caption{Mean absolute error of PhysNet models that were pretrained on UBFC then test-time adapted to DDPM as a function of $N_{TTA}$.
    }
    \label{fig:TTA_over_updates}
\end{figure}

In Sec.~\ref{sec:personalization} we described a scenario where the model could adapt itself to a new subject by training on the first 20 seconds of video from that person.
The general trend of our experiments has moved from large-scale unsupervised training towards data-scarce settings where models become specialized for the testing domain.
The next step is to ask whether a model can adapt to a single new testing sample.
In this section, we describe experiments in test-time adaptation, where a model is continuously trained on incoming samples.
Figure~\ref{fig:TTA_and_Personalization} shows an overview of test-time adaptation with SiNC.

There are known benefits and risks of such an approach.
The benefit is that the model can theoretically adjust to domain shifts in the input data quickly.
The downside is that the model can undergo catastrophic forgetting~\cite{Goodfellow2014}, where it can no longer generalize to data in the domain upon which it was originally trained.
Furthermore, inference time for the adapted model increases from a single forward pass to the number of training iterations times the sum of a forward and backward pass.
However, if few training iterations are required and the adaptation is rapid, catastrophic forgetting no longer becomes an issue.
Rather, the inference process for the approach becomes the gradient-based updates to the model followed by a forward pass on the same sample.
This approach is substantially simpler than~\cite{Lee_ECCV_2020}, which requires another pretrained network to generate gradients for adapting the model weights.

Differently from Sec.~\ref{sec:cross_dataset}, we perform our experiments with pretrained models from UBFC, PURE, and DDPM, and attempt to quickly adapt to incoming cross-dataset samples.
For example, each of the subjects in UBFC has 15 adapted models that were originally trained on PURE.
We use the horizontal flipping, illumination noise, random cropping, and time reversal augmentations to create a batch of 20 samples from a single 120-frame input clip.
We sequentially process the video with a 60-frame stride and update the model weights $N_{TTA}$ iterations for each new testing clip, then make a prediction on the unperturbed clip with the adapted model.

\begin{figure}
    \centering
    \includegraphics[width=\linewidth]{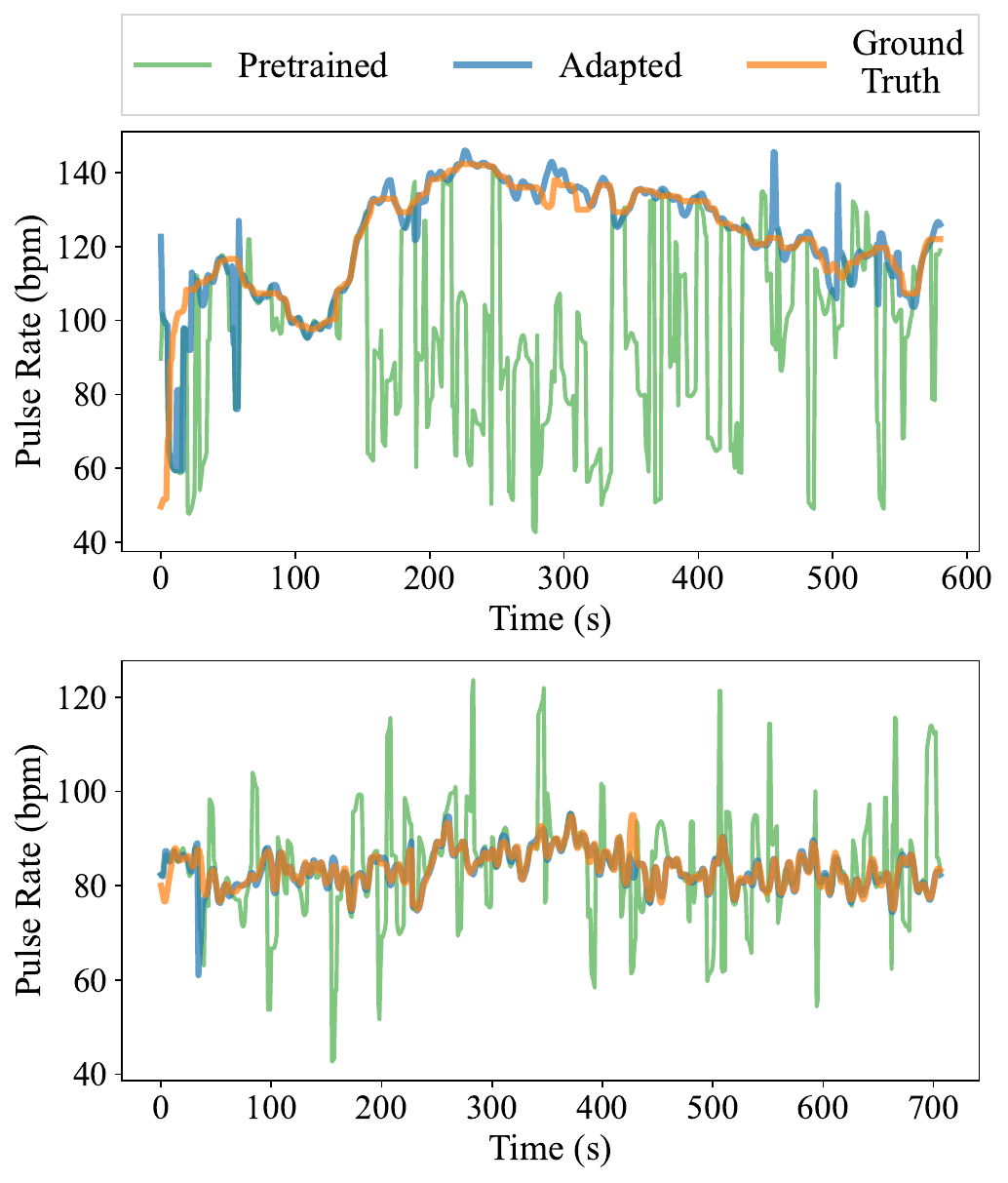}
    \caption{Pulse rate estimates of two DDPM samples with baseline and test-time adapted PhysNet models that were pretrained on the UBFC dataset. Models were adapted $N_{TTA}=1$ times on each test sample, so predictions tend to improve as the model adapts over time.}
    \label{fig:TTA}
\end{figure}

We find this simple test-time adaptation strategy to be extremely effective in handling domain shifts.
Table~\ref{tab:cross_dataset} shows the cross-dataset performance when using test-time adaptation with $N_{TTA}=1$, labeled as ``SiNC+TTA''.
Using test-time adaptation consistently gives the best results, even though these models have never been trained with a ground truth PPG signal.
Adapted PhysNet models that were pretrained on PURE and UBFC perform better than training a model from scratch on all of the DDPM training set.
Figure~\ref{fig:TTA} shows example predictions from fixed and adapted UBFC models tested on DDPM.

Test-time adaptation comes with the downside of increased runtime.
However, when these operations were performed on a GPU, test-time adaptation with $N_{TTA} \leq 2$ was faster than real-time.
We performed an ablation study on the $N_{TTA}$ hyperparameter, which determines the number of model weight updates for each incoming test sample.
Figure~\ref{fig:TTA_over_updates} shows the MAE when adapting UBFC models to DDPM for different values of $N_{TTA}$.
Intuitively, when updating a model less than once every incoming sample the model does not adapt quickly enough and performs poorly (e.g. $N_{TTA}=\frac{1}{25}$).
Around $N_{TTA}=\frac{1}{2}$, the model matches within-dataset performance, and when $N_{TTA} \in \{1,2\}$ the adapted model outperforms within-dataset models.
If $N_{TTA}$ is too large, the training dynamics become unpredictable and the model may overfit to each incoming sample.
For simplicity, we recommend setting $N_{TTA}=1$, where a single weight update is applied on each new sample.
\section{Poisoned or Noisy Training Data}
\label{sec:poisoned}

In previous experiments we ignored the ground truth PPG signals, but we trained with videos from baseline rPPG datasets. The videos from these datasets tend to be high quality and the subjects actions are constrained for easy pulse estimation. However, in practice, collecting large unlabelled video datasets would result in many ``negative" samples which are unusable for rPPG. For example, some videos may be too compressed, contain too few skin pixels, have strong motion, or insufficient lighting. 

This motivates experiments on the robustness of SiNC during training with unusable training samples.
We conduct an experiment similar to datset poisoning for unsupervised approaches~\cite{Carlini2021}.
First, we create negative samples that do not contain a pulse, but visually look similar to other training samples.
Specifically, add dynamic Gaussian noise over time to a constant frame of the face, as was done in~\cite{Speth2023HallucinatedHA}.
The negative video clips are sampled and shown to the model during training with a probability of $\alpha$.
We run multiple training iterations varying $\alpha$ to see how the poisoning percentage affects model convergence. 

\begin{figure}
    \centering
    \includegraphics[width=\linewidth]{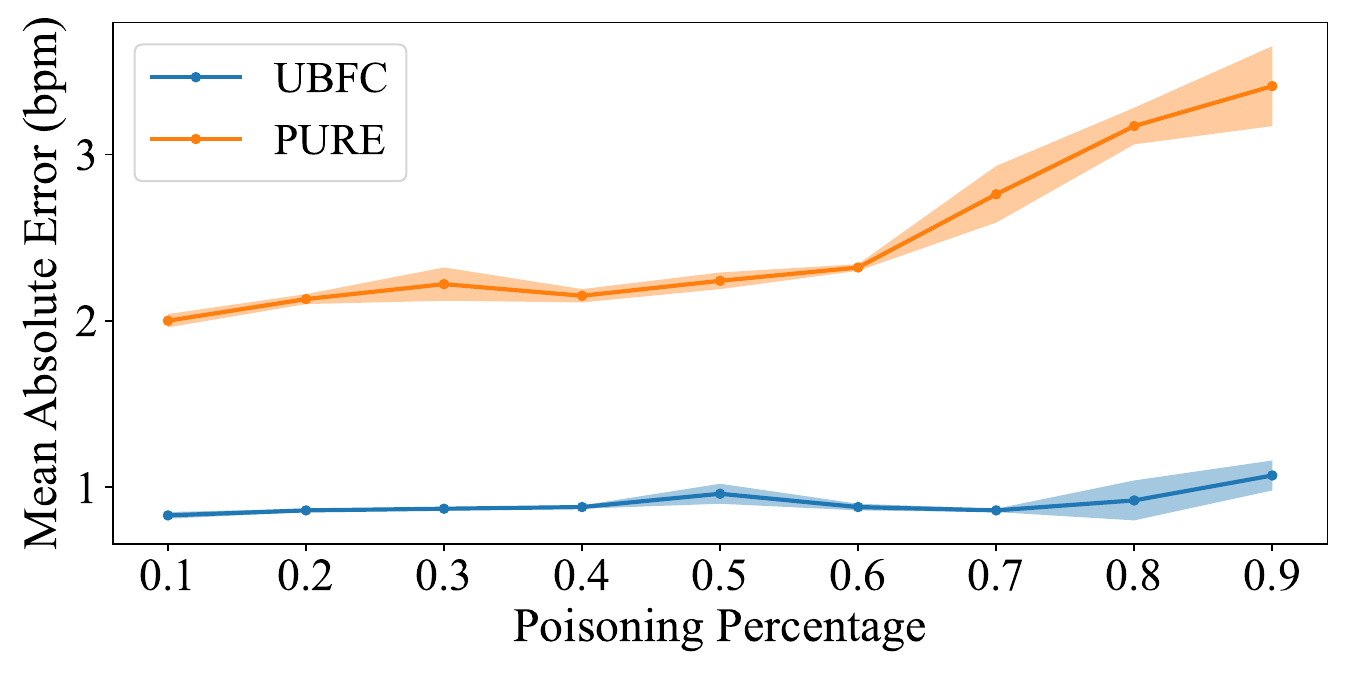}
    \caption{Mean absolute error of TSCAN models trained on UBFC with different rates of poisoning.}
    \label{fig:poisoning}
\end{figure}

For our experiments, we trained the TSCAN model with SiNC on poisoned versions of the UBFC~\cite{Bobbia2019} dataset. We used a 5-fold cross validation strategy similarly to ~\cite{speth2023sinc}, and trained 3 models for each fold to examine variation across network initializations. Figure~\ref{fig:poisoning} shows the mean absolute error (MAE) on UBFC and PURE~\cite{Stricker2014} as a function of the poisoning percentage. In general, SiNC appears relatively robust to dataset poisoning, giving less than 1 bpm of MAE degradation for within-dataset performance even up to 90\% poisoning. For cross-dataset performance, there appears to be a critical change near 60\% poisoning, where the MAE starts to increase rapidly. Overall, these experiments show that SiNC only requires a small percentage of high quality training samples to converge to signal of interest.

\section{Camera-Based Respiration Estimation}
\label{sec:respiration}

Several useful health diagnostics can be inferred from a subject's breath~\cite{Nicolo2020}. As remote vitals systems improve, the most powerful systems will extract both rPPG and respiration simultaneously to better understand one's health.
To show that SiNC is a general framework for learning periodic signals, we apply it to the problem of remote respiration estimation. Respiration is different from rPPG in many ways.
The primary difference is that the visual dynamics of respiration are carried in motion rather than chromatic features.
Secondly, the physiological limits for the breathing rate are different than rPPG, so it tests whether the loss formulations for SiNC are general enough for a different periodic signal. 

One of the strengths of SiNC is the limited number of hyperparameters needed for training. 
The only parameters that need to be adjusted for learning respiration are the bandlimits for the losses, $a$ and $b$, and the width of the signal band, $\Delta_F$, in the sparsity loss.
For respiration training, we set the low cutoff, $a$, as 0.1 Hz (6 breaths per minute) and the high cutoff, $b$ as 0.5 Hz (30 breaths per minute). Although rates near these limits are unlikely to be observed in practice, we found that they were still narrow enough for the model to learn the respiration signal easily.
We set $\Delta_F$ to $0.02\Bar{66}$ Hz, which is proportional to the area under the main lobe in our rPPG experiments.

 \begin{figure}
   \centering
   \includegraphics[width=\columnwidth]{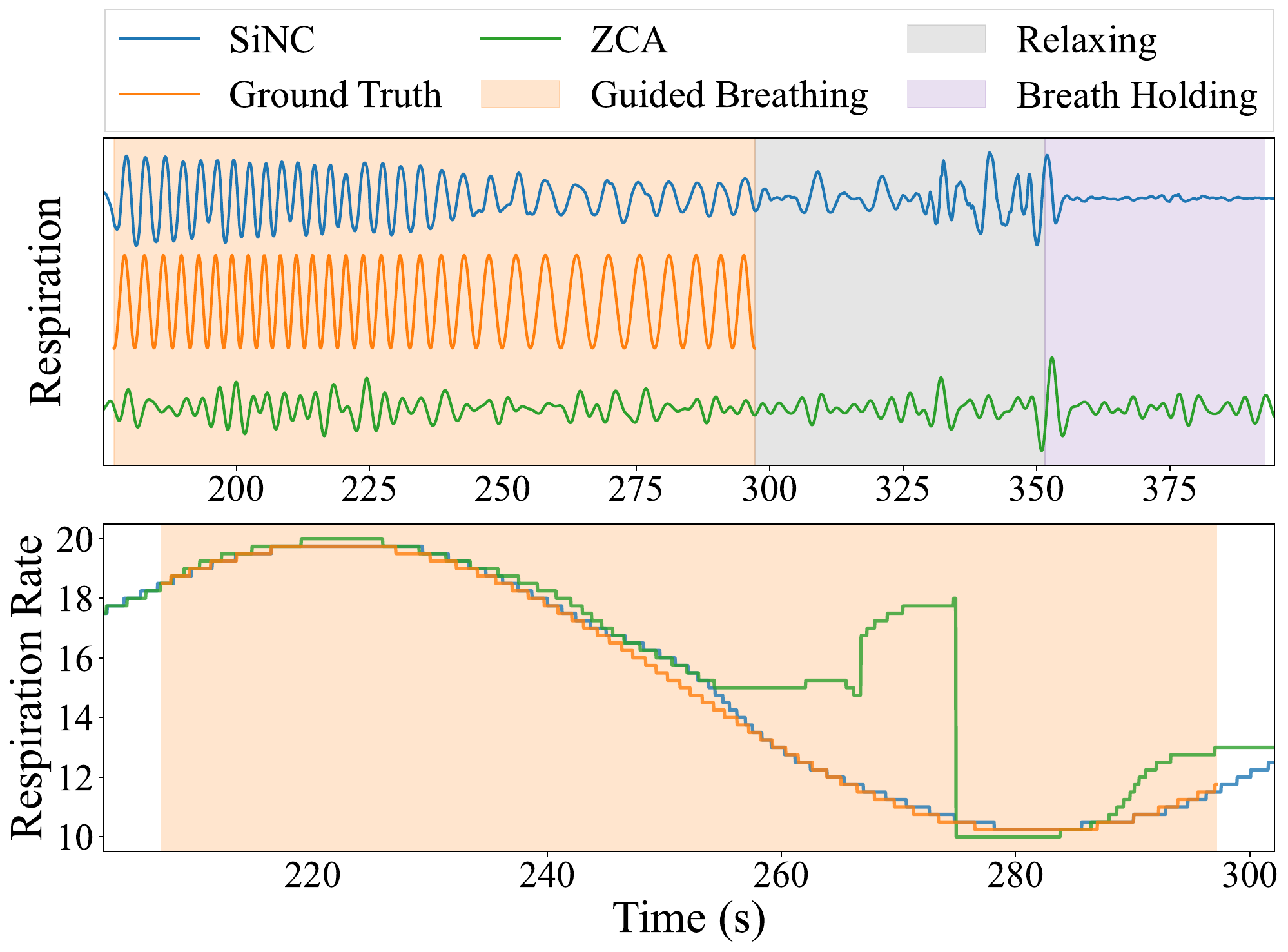}
   \caption{Remote respiration predictions on the MSPM dataset for a PhysNet model trained with SiNC on RGB video at 10 fps. Respiration frequency is accurate during the guided breathing activity, and shows low amplitude ``flatlining" during the breath holding activity.}
   \label{fig:respiration}
 \end{figure}

\subsection{Respiration Video Preprocessing}
We use the MSPM dataset~\cite{Speth_TIM_2023} for our experiments, since they contain 2 minutes with respiration labels for evaluation.
To prepare the input video clips for respiration estimation, we crop the chest region rather than the face.
We first selected the minimum and maximum (x,y) locations of the ``upper-clothes" mask from the human parsing package~\cite{li2020self}.
A static bounding box was created from the mean of these (x,y) coordinates for the entire video.
The static bounding box is necessary to avoid introducing motion.
An example of the preprocessing for respiration is shown on the left side of Fig.~\ref{fig:preprocessing}.
Since respiration is usually carried in the motion signal, for a subset of experiments we applied dense optical flow~\cite{FarnebackOF} to test if optical flow inputs are easier to learn from than RGB inputs.
For respiration experiments we use an input sample duration of $T=300$ frames (10 seconds), since it is lower frequency than pulse.

We conducted experiments with video at 30 fps and downsampled versions at 10 fps, since the sampling rate for respiration could theoretically go down to 1 fps with the selected upper cutoff of 0.5 Hz.
We trained both PhysNet~\cite{Yu2019} and TSCAN~\cite{Liu_MTTS_2020} model architectures for the respiration task at both frame rates for thorough evaluation.
The PhysNet models took downscaled $64 \times 64$ pixel inputs, while the TSCAN model took $36 \times 36$ pixel inputs. We trained PhysNet models with either direct RGB input or 2 channel optical flow input. Models were trained for 40 epochs.

\subsection{Respiration Results}
As a baseline comparison for remote respiration we applied the best-performing approach from ~\cite{Zhan2020Resp}. This algorithm uses the dense optical flow as input. The vertical (y-axis) motion was used, and the horizontal (x-axis) movement was discarded.
The dense optical flow videos were then downscaled to $10 \times 10$ pixels, flattened to an $N \times 100$ matrix, then we performed zero-phase component analysis (ZCA).
From all of the projected components, we averaged the 3 with the highest signal-to-noise ratio (SNR).
The SNR was calculated as the sum of signal power between 10 and 20 breaths per minute, divided by the power outside those bounds.

\begin{table}\footnotesize
    \centering
    \caption{Remote respiration results. ``RGB" denotes raw pixels as input, and ``OF" denotes dense optical flow as input.}
    \setlength{\tabcolsep}{3.0pt}
    \resizebox{\columnwidth}{!}{%
    \begin{tabular}{lccccc}
        \toprule
        Method & Input & FPS & MAE & RMSE & r \\
        \midrule
        ZCA~\cite{Zhan2020Resp} & OF & 30 & 1.14 & 2.44 & 0.79 \\
        SiNC (PhysNet) & OF & 10 & 0.24 $\pm$ 0.02 & 0.80 $\pm$ 0.10 & 0.98 $\pm$ 0.01 \\
        SiNC (PhysNet) & OF & 30 & \bf 0.21 $\pm$ 0.00 & \bf 0.66 $\pm$ 0.03 & \bf 0.99 $\pm$ 0.00 \\
        SiNC (PhysNet) & RGB & 10 & 0.32 $\pm$ 0.01 & 1.13 $\pm$ 0.03 & 0.96 $\pm$ 0.00 \\
        SiNC (PhysNet) & RGB & 30 & 0.77 $\pm$ 0.08 & 2.05 $\pm$ 0.17 & 0.85 $\pm$ 0.02 \\
        SiNC (TSCAN) & RGB & 10 & 0.36 $\pm$ 0.09 & 1.19 $\pm$ 0.24 & 0.95 $\pm$ 0.02 \\
        SiNC (TSCAN) & RGB & 30 & 0.83 $\pm$ 0.06 & 2.02 $\pm$ 0.14 & 0.86 $\pm$ 0.02 \\
        
        \bottomrule
    \end{tabular}}
    \label{tab:respiration}
\end{table}


We evaluated the different methods on respiration rate estimation by using a sliding 30-second STFT window over the predicted respiration time signal. Table~\ref{tab:respiration} shows the results over the whole MSPM dataset, where MAE and RMSE are presented in units of breaths per minute. In general, we find that the PhysNet models trained with SiNC perform best with correlations of 0.97 and 0.92, for models given dense optical flow and RGB as input, respectively. As we expected, dense optical flow provides direct input features and makes the signal easier to learn for the model.
Figure \ref{fig:respiration} shows the ground truth respiration waveform and frequencies alongside the camera-based respiration estimated from a video.
During the ``Relaxing'' activity the signal quality degrades due to subjects recovering from the guided breathing activity and talking.
During the ``Breath-Holding'' activity the model gives a low amplitude signal, which could be promising for detecting apnea.

\section{Discussion}
\subsection{Improvements over Supervised Learning}
It is initially surprising that unsupervised training leads to similar or improved rPPG estimation models compared to those trained in a supervised manner. However, there are several potential benefits to unsupervised training. From a hardware perspective, one of the difficulties in supervised training is aligning the contact pulse waveform with the video frames~\cite{Zhan2020}. The pulse sensor and camera may have a time lag, effectively giving the model an out-of-phase target at training time. Unsupervised training gives the model freedom to learn the phase directly from the video. The contact-PPG signal is also sensitive to motion and may be noisy. Since motion may co-occur at the face and fingertip, the contact signal may misguide the model towards visual features for which they should be invariant.

From a physiological perspective, the pulse observed optically at the fingertip with a contact sensor has a different phase than that of the face, since blood propagates along a different path before reaching the peripheral microvasculature, making alignment nearly impossible without shifting the targets to rPPG estimates from existing methods~\cite{Speth_CVIU_2021}. Additionally, the morphological shape of the contact-PPG waveform depends on numerous factors such as the wavelength of light (and corresponding tissue penetration depth), external pressure from the oximeter clip, and vasodilation at the measurement site~\cite{Moco2018,Abraham2013}. This indicates that the morphology and phase of the target PPG waveform is likely different from the observed rPPG waveform.

\subsection{Why Does It Work?}
The success of \ourapproach depends on specific properties of the data, model, and how the two interact.
Limited model capacity is actually required, since it forces learning features that generalize across inputs.
By constraining the model's predictions to have specific periodic properties, the limited-capacity model must find a general set of features to produce a signal that exists in most or all of the training samples, which is the blood volume pulse or respiration in our datasets.

As a beneficial side-effect, the model intrinsically learns to ignore common noise factors such as illumination, rigid motion, non-rigid motion (\eg talking, smiling, etc.), and sensor noise, since they may preside outside the predefined bandlimits or with uniform power spectra.
Even if noise exhibits periodic tendencies within the bandlimits for some samples, those features would produce poor signals on other samples.
Therefore, end-to-end unsupervised approaches like \ourapproach are well-suited for periodic problems.

\section{Conclusions}
We proposed a novel non-contrastive learning approach for end-to-end unsupervised signal regression, with specific experiments on blood volume pulse and respiration estimation from videos.
This SiNC framework effectively learns powerful visual features with only loose frequency constraints.
We demonstrated this by training accurate rPPG models using non-rPPG data and our simple loss functions.
We also showed that the SiNC framework is effective with very little training data.
Our experiments demonstrated that SiNC can be used for both single-subject personalization and test-time adaptation, allowing models to quickly update their weights to handle domain shifts.
Given the subtlety of the rPPG and respiration signals, we believe our work can be extended to other signal regression tasks in domains outside of remote vitals estimation.

\section*{Acknowledgments}

The work presented in Sections \ref{sec:intro} through \ref{sec:personalization} was sponsored by the Securiport Global Innovation Cell, a division of Securiport LLC. Commercial equipment is identified in this work in order to adequately specify or describe the subject matter. In no case does such identification imply recommendation or endorsement by Securiport LLC, nor does it imply that the equipment identified is necessarily the best available for this purpose. The opinions, findings, and conclusions or recommendations expressed in this publication are those of the authors and do not necessarily reflect the views of our sponsors.

The work presented in Sections \ref{sec:adaptation}, \ref{sec:poisoned} and \ref{sec:respiration} was supported by the U.S. Department of Defense (Contract No. W52P1J2093009). The views and conclusions contained in this document are those of the authors and should not be interpreted as representing the official policies, either expressed or implied, of the U.S. Department of Defense or the U.S. Government. The U.S. Government is authorized to reproduce and distribute reprints for Government purposes, notwithstanding any copyright notation here on.

\bibliographystyle{IEEETran}
\bibliography{biblio}

\end{document}